\documentclass[sigconf]{acmart}

\AtBeginDocument{%
  }

\copyrightyear{2023} 
\acmYear{2023} 
\setcopyright{rightsretained} 
\acmConference[ICMR '23]{International Conference on Multimedia Retrieval}{June 12--15, 2023}{Thessaloniki, Greece}
\acmBooktitle{International Conference on Multimedia Retrieval (ICMR '23), June 12--15, 2023, Thessaloniki, Greece}\acmDOI{10.1145/3591106.3592230}
\acmISBN{979-8-4007-0178-8/23/06}

\usepackage{colortbl}

\usepackage{makecell}
\usepackage{multirow}
\usepackage{tabularx}

\usepackage{longtable} 
\usepackage{hyperref} 
\usepackage{arydshln} 

\newcolumntype{Y}{>{\centering\arraybackslash}X}

\begin{document}
%
\title{Improving Generalization for Multimodal Fake News Detection}



\author{Sahar Tahmasebi}
\email{sahar.tahmasebi@tib.eu}
\orcid{0000-0003-4784-7391}
\affiliation{%
  \institution{TIB – Leibniz Information Centre for Science and Technology}
  \city{Hannover}
  \country{Germany}
}

\author{Sherzod Hakimov}
\email{sherzod.hakimov@uni-potsdam.de}
\orcid{0000-0002-7421-6213}
\affiliation{%
  \institution{University of Potsdam}
  \city{Potsdam}
  \country{Germany}
}


\author{Ralph Ewerth}
\authornote{Also affiliated with TIB -- Leibniz Information Centre for Science and Technology, Hannover, Germany}
\orcid{0000-0003-0918-6297}
\email{ralph.ewerth@tib.eu}
\author{Eric Müller-Budack}
\authornotemark[1]
\orcid{0000-0002-6802-1241}
\email{eric.mueller@tib.eu}
\affiliation{%
  \institution{
  L3S Research Center, Leibniz University Hannover
  }
  \city{Hannover}
  \country{Germany}
}




\begin{abstract}

The increasing proliferation of misinformation and its alarming impact have motivated both industry and academia to develop approaches for fake news detection. 
However, state-of-the-art approaches are usually trained on datasets of smaller size or with a limited set of specific topics. 
As a consequence, these models lack generalization capabilities and are not applicable to real-world data. 
In this paper, we propose three models that adopt and fine-tune state-of-the-art multimodal transformers for multimodal fake news detection.
We conduct an in-depth analysis by manipulating the input data aimed to explore models performance in realistic use cases on social media. Our study across multiple models demonstrates that these systems suffer significant performance drops against manipulated data.
To reduce the bias and improve model generalization, we suggest training data augmentation to conduct more meaningful experiments for fake news detection on social media. 
%
The proposed data augmentation techniques enable models to generalize better and yield improved state-of-the-art results.

%

%
%
%
%
\end{abstract}

\ccsdesc[500]{Information systems~Multimedia information systems}
\ccsdesc[300]{Information systems}


\keywords{Multimodal fake news detection, 
social media, news analytics}



\maketitle

\section{Introduction}

Misinformation and fake news that contain false information to deliberately deceive readers~\cite{allcott2017social}, have become a pressing challenge for society. 
Typically, fake news use different modalities such as images, text, and videos to attract attention. 
Thus, automated \textit{multimodal} solutions that incorporate information from multiple modalities are essential to detect misinformation on social media or in news.

%
Multimodal research on fake news detection has been growing rapidly, and current approaches~\cite{DBLP:att-RNN,DBLP:Spotfake,DBLP:BDANN,DBLP:EANN} have mainly used deep learning to extract features from text and image.
For example, Wang et al.~\cite{DBLP:EANN} introduced an \textit{Event Adversarial Neural Network~(EANN)}, which extracts event-invariant features to detect fake news on unforeseen events. 
\textit{Spotfake}~\cite{DBLP:Spotfake} 
demonstrated the effectiveness of combining pre-trained language models~\cite{DBLP:BERT} and computer vision approaches~\cite{DBLP:VGG19}.
%
Zhang et al.~\cite{DBLP:BDANN} added a domain classifier and proposed a \textit{BERT}-based~(Bidirectional Encoder Representations from Transformers) domain adaptation neural network~(\textit{BDANN}) for multimodal fake news detection.
These approaches achieve impressive results but are trained on rather small datasets with similar characteristics between train and test data.
However, due to the dynamic nature of news and realistic use cases for social media, where the news could be affected by human-produced false content, they mostly fail in real-world scenarios. 
This failure mostly arises from overfitting to input data resulting in a lack of generalization. 

%
In this paper, we address the aforementioned issues and present the following contributions.
(1)~We propose three multimodal models that adapt state-of-the-art
transformers for fake news detection in social media. 
Experimental results on the \textit{MediaEval Benchmark} for the \textit{Verifying Multimedia Use} task~\cite{DBLP:MediaEval2015,MEdiaEval2016} demonstrate the effectiveness of these models compared to the state of the art. 
(2)~We suggest several content manipulation techniques to provide new settings for training and evaluation and conduct an in-depth analysis of model performance in more meaningful test scenarios 
that confirm poor generalization of previous approaches.
%
(3)~We suggest training data augmentation using the large-scale \textit{Visual News} dataset~\cite{DBLP:VN} that contain news from a wide range of topics to improve generalization. 
(4)~Furthermore, we introduce an ensemble approach that combines models trained with various manipulation techniques to further improve generalization. Experimental results have demonstrated that it provides the best overall results across various test sets. 
Source code, models, and dataset are publicly made available on \textit{GitHub}\footnote{\textit{GitHub} repository: \url{https://github.com/TIBHannover/MM-FakeNews-Detection}\label{foot:github}} to allow for fair comparisons.

The remainder of this paper is organized as follows. 
The multimodal models and data augmentation for fake news detection are proposed in Section~\ref{sec:methodology}. 
The experimental setup, results and a revised training setup to improve generalization are presented in Section~\ref{sec:experiments}. Section~\ref{sec:conclusion} concludes this paper and provides future work directions.


\section{Multimodal Fake News Detection}
\label{sec:methodology}
In this section, we propose data augmentation strategies to improve model generalization as well as three models based on recent transformers for multimodal fake news detection.
\begin{table}
    \centering
    \caption{Distribution of training, validation, and test samples.} 
    \label{tab:ME_dataset}
    \small
    \begin{tabularx}{1.0\linewidth}{Xccccc}
    \toprule
    \bf{Dataset } & 
    \bf{Category} & 
    \bf{Train} & 
    \bf{Validation} &
    \bf{Test} &
    \bf{All}
    \\
    \midrule

    \multirow{2}{*}
    {\makecell{MediaEval\\2015~\cite{DBLP:MediaEval2015}}}    & Fake &    5,245   &   1,479   &  717 &   \multirow{2}{*}{13,480}\\
    & Real & 3,426   &    1,398   &  1,215 & \\
    \midrule
    \multirow{2}{*}{\makecell{MediaEval\\2016~\cite{MEdiaEval2016}}}    & Fake &    5,962   & 1,479   &   630 & \multirow{2}{*}{14,578}\\
    & Real & 4,641 &   1,398   &   468 &\\
    \bottomrule
\end{tabularx}
\end{table}
\subsection{Dataset Augmentation}
\label{sec:augmentation}
As mentioned before, datasets for multimodal fake news detection~\cite{DBLP:MediaEval2015, MEdiaEval2016} are typically rather small and contain limited topics. As a consequence, supervised models tend to overfit which limits their performance on unseen data.  
While large-scale news dataset that cover a huge variety of topics are available, they usually contain only \textit{real} articles and are not labeled for multimodal fake news detection.
We propose to use various content manipulation techniques, such as the replacement of entities~(e.g., events) or images, to automatically create \textit{fake} samples to train deep learning approaches for multimodal fake news detection with better generalization capabilities. Details on the manipulations techniques and training strategies including an ensemble model are provided in Section~\ref{sec:result_analysis}.
\subsection{Multimodal Transformers Models}
\label{sec:transformers}
We propose three multimodal approaches for fake news detection to predict whether a post is \textit{fake}~($y=0$) or \textit{real}~($y=1$). 
They consist of three modules: (1)~A modal-specific encoder that extracts multimodal features from both text~$\mathbf{x}_{T}$ and image~$\mathbf{x}_{I}$, 
(2)~a fusion module that combines the features from both modalities, and  
(3)~a classification module that outputs class probabilities. 
The goal is to find the best model~$\mathcal{F}(\mathbf{x}_{T}, \mathbf{x}_{I})\,\to\ y \in {\{0,1}\}$.

\subsubsection{BERT-ResNet Model}
\label{sec:bertresnet}
This model adopts pre-training techniques to encode each modality. For the textual part~$T$, it uses 
\textit{BERT}~\cite{DBLP:BERT} to obtain text embedding~$\mathbf{x}_{T}$ as it achieves good results for various natural language processing tasks.
We use a residual network with 50 layers~(\textit{ResNet-50})~\cite{DBLP:ResNet} pre-trained on \textit{ImageNet}~\cite{Russakovsky2015} to extract image representations~$\mathbf{x}_{I}$.
By using two fully-connected~(FC) layers of size 256 and \textit{GELU~(Gaussian Error Linear Unit)}~\cite{GLU} activation, we transform the embeddings of both modalities to a joint space ($\mathbf{x}_{T},\mathbf{x}_{I}\in\mathbb{R}^{256}$). 
The embeddings are concatenated and passed to a FC layer with size 128 and \textit{ReLU~(Rectified Linear Unit)} activation. Finally, a FC classification layer with two neurons and \textit{sigmoid} activation function is used to output probabilities for \textit{fake} and \textit{real}.

\subsubsection{MLP-CLIP Model}
This model extracts multimodal features using the \textit{ViT-B/32} variant of \textit{CLIP}~(contrastive language-image pre-training)~\cite{DBLP:CLIP}, which has proven to be powerful for many downstream tasks.
%
%
The \textit{CLIP} model outputs two embeddings of the same size for image~$\mathbf{x}_{I}$ and text~$\mathbf{x}_{T}$. 
Each embedding is fed into an multi-layer perceptron~(\textit{MLP}) comprising two FC layers with sizes 256 and 128 and \textit{ReLU} activation function. 
We concatenate the embeddings and feed them into a FC layer with \textit{sigmoid} activation and two neurons to output class probabilities for \textit{fake} and \textit{real}.

\subsubsection{CLIP-MMBT Model}
This approach leverages a multimodal fusion method called \textit{MultiModal BiTransformers~(MMBT)}~\cite{DBLP:MMBT} for classification. 
%
Pre-trained \textit{ResNet-50x4} variant of \textit{CLIP}~\cite{DBLP:CLIP} and \textit{BERT}~\cite{DBLP:BERT} are used to create representation for image~$\mathbf{x}_{I}$ and text~$\mathbf{x}_{T}$, respectively.
Both representations are passed through \textit{MMBT} for multimodal fusion that uses a FC layer with one output neuron and \textit{sigmoid} activation as the last layer to output a probability~$\rho \in [0, 1]$ that is used to predict the class \textit{fake}~($\rho < 0.5$) or \textit{real}~($\rho \geq 0.5$).
\begin{figure*}[t]
 \centering
 \includegraphics[width=0.88\linewidth]{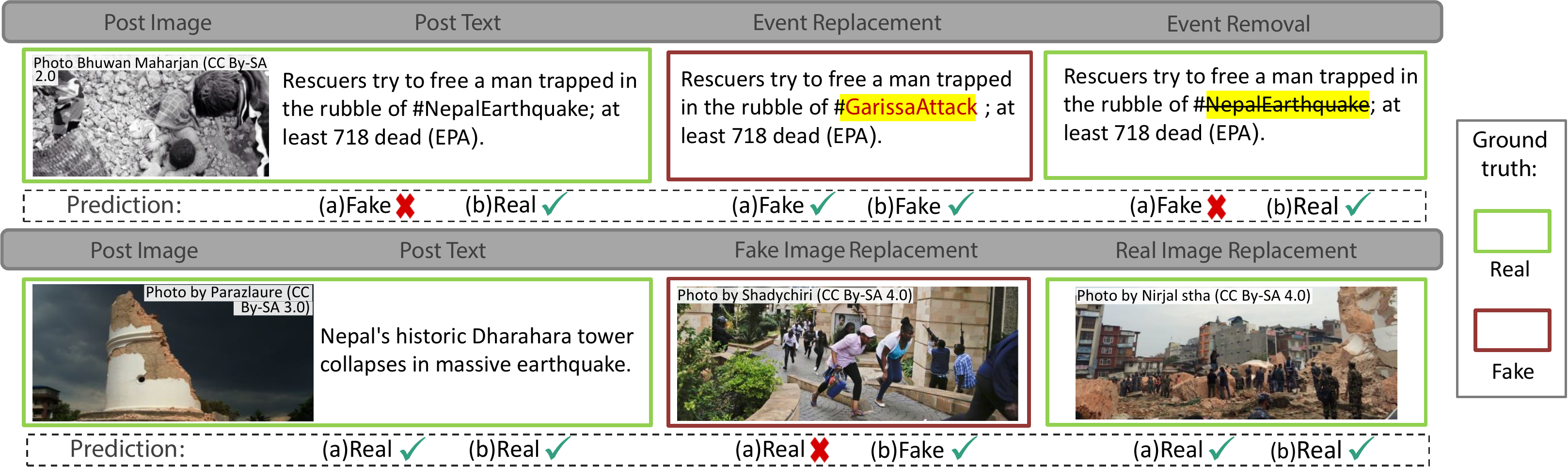} 
 \caption{Manipulation techniques and results of (a)~Spotfake and (b)~MLP-CLIP~(VNME-Ens). Images are replaced with similar ones due to licensing issues. Links to the original images are provided on \textit{GitHub}\textsuperscript{\ref{foot:github}}.
 }
 \label{Fig: Fig1}
\end{figure*}

\section{Experimental Setup and Results}

\label{sec:experiments}
%
\begin{table}[t]
    \small
    \centering
    \caption{Precision~(P), recall~(R), and F1-score~(F1) of our models and the state of the art. Results in gray serve as reference but are \textit{not directly comparable} due to different experimental setups~(e.g. train and test splits).
     Reproduced results using official implementations are denoted with * and are directly comparable. Results in bold are the best on our test splits.} 
    
    \label{table:performances}
\setlength{\tabcolsep}{2pt}

\begin{tabularx}{1.0\linewidth}{lYYYYYY}
    \toprule
    &
    \multicolumn{3}{c}{\bf{MediaEval (ME) 2015}} &
    \multicolumn{3}{c}{\bf{MediaEval (ME) 2016}} \\

    \cmidrule(lr){2-4}
    \cmidrule(lr){5-7}
    \bf{ } & 
    \bf{P} & 
    \bf{R} & 
    \bf{F1} &
    \bf{P} &
    \bf{R} &
    \bf{F1}
    \\
    \midrule
    \rowcolor[gray]{.9}
    SpotFake \cite{DBLP:Spotfake}   &  -   &   -   &   -   & 0.791 &  0.753 & 0.760   \\
    \rowcolor[gray]{.9}
    MVAE \cite{DBLP:MVAE}  &   -   &  -   &  -  &  0.745   & 0.748 &   0.744 \\
    \rowcolor[gray]{.9}
    BDANN \cite{DBLP:BDANN}   &   -   &   -   &   -   &   0.820   &   0.780
    & 0.795\\
    \rowcolor[gray]{.9}
    Singhal et al. \cite{DBLP:Singhal}   &   -   &   -   &   -   &   0.836   &   0.832 &   0.830\\
    \rowcolor[gray]{.9}
    EANN \cite{DBLP:EANN}   &   0.847   &   0.812 &   0.829 &   -   &   -   & -\\
    \rowcolor[gray]{.9}
    Singh et al \cite{singh}  &   0.867   &   0.844   &   0.848   &   -   & -   &   -\\
    \rowcolor[gray]{.9}
    CAFE \cite{DBLP:CAFE}    &   0.806   &   0.806   &   0.806   &   -   &   - &   -\\

    \hdashline
    Spotfake*   &   0.831   &   0.824   &   0.827   &   0.751 &   0.723   &   0.727\\
    BDANN*  &   0.780   &   0.711   &   0.723   &   0.700   & 0.652   &   0.599\\
    \midrule
    BERT-ResNet &   0.870   &   0.873   &   0.872   &   0.634   &   0.631   &   0.632\\
    MLP-CLIP    &   \bf{0.926}   &   \bf{0.943}   &   \bf{0.933}   &   \bf{0.807}   &   \bf{0.806}  &   \bf{0.806}\\
    CLIP-MMBT   &   0.860   &   0.673   &   0.677   &   0.689   &   0.644   &   0.640\\

    
    \bottomrule
\end{tabularx}

\end{table}

This section provides details on the experimental setting~(Section~\ref{sec:params}), a comparison to the state of the art~(Section~\ref{sec:sota_comparison}), and impact of content manipulation including result analysis  
(Section~\ref{sec:result_analysis}). 
\subsection{Experimental Setting}
\label{sec:params}
\subsubsection{Implementation details:}
We tried multiple hyperparameters for all models and provide the best setups and models on \textit{GitHub}\textsuperscript{\ref{foot:github}}. 
All models are trained for 30 epochs with a batch size of 32 and the best model with the lowest validation loss is used for testing.
\subsubsection{Datasets:}
We use the \textit{2015} and \textit{2016} \textit{MediaEval Benchmark} for the \textit{Verifying Multimedia Use} task~\cite{DBLP:MediaEval2015,MEdiaEval2016}. 
To the best of our knowledge, they are the only publicly available datasets for fake news detection containing both image and text of the tweets written in English. 
Both datasets have a development and test set, each with its own events. 
Each tweet contains text with corresponding images or video. Given our focus on image content, we remove tweets with videos and tweets for which associated images are not available anymore. 
Table~\ref{tab:ME_dataset} shows the dataset statistics. Data splits are provided on \textit{GitHub}\textsuperscript{\ref{foot:github}} to allow for fair comparisons.
\subsection{Comparison to the State of the Art}
\label{sec:sota_comparison}
Table~\ref{table:performances} presents a comparison of our proposed models with the state of the art.  
%
Although related work has been evaluated on the same dataset, different sizes of subsets for training, validation, and test were used.
Reasons are that there are some samples whose images are no longer available or specific data filtering methods~(e.g., based on text length). 
As a consequence, results are reported with different experimental setups and \textit{not directly comparable}. 
Nevertheless, we include results from related work for reference. 

To ensure a fair evaluation, we reproduce results for \textit{BDANN}~\cite{DBLP:BDANN} 
and \textit{Spotfake}~\cite{DBLP:Spotfake}
since they provide applicable official \textit{GitHub} implementations. 
We choose the best hyperparameters reported in their paper.
The \textit{MLP-CLIP model} achieves the best results among our models and the reproduced models from the state of the art for both datasets. 
This suggests that 
the \textit{CLIP model} learns expressive visual and textual representations, which can be used in multimodal fake news detection. 
%
%
%
Table~\ref{table:performances} also shows that the reproduced results for \textit{Spotfake} and \textit{BDANN} are worse compared to the reported ones~(gray background) that have been obtained with a different experimental setup~(e.g., train and test split).
Their lower performance implies that our experimental setup is more challenging. 
\subsection{Impact of Content Manipulation}
\label{sec:result_analysis}
Despite good test performances, model generalization is questionable due to the scarcity of real events and limited number of unique images in the \textit{ME} datasets. 
This can lead classifiers to learn domain-specific features instead of making an actual distinction between \textit{fake} and \textit{real}.
In this section, we analyze fake news detection models in various use cases with manipulated samples. 
As proposed in Section~\ref{sec:augmentation}, we apply data augmentation to avoid overfitting during model training 
to improve generalization.
\begin{table}[t]
\caption{Number of train, validation~(valid.), and test samples for different manipulation strategies~(MS) used to compose different versions of the VNME dataset.}
    \label{tab:VN_data}
    \small
    \setlength{\tabcolsep}{4.5pt}
    \begin{tabularx}{1.0\linewidth}{Xccccccc}
    \toprule
    \multirow{2}{*}{\bf{Data}}&
    \multirow{2}{*}{\bf{Split}}&
    \multirow{2}{*}{\bf{MS}}&
    \multirow{2}{*}{\bf{Fake}}& 
    \multirow{2}{*}{\bf{Real}}&
    \multicolumn{3}{c}{\bf{VNME}} \\
    &&&&&\bf{(Img)}&\bf{(Evt)}&\bf{(All)}\\
    \midrule
    \multirow{6}{*}{\makecell{\textbf{Visual}\\\textbf{News}\\\textbf{(VN)}}}& Train &   -   &   0&258,488   &   \checkmark   &   \checkmark    &   \checkmark  \\
    &Train &   EvRep&245,384&  0  &   $\times$   &   \checkmark    &   \checkmark \\
    &Train &   FakeIm  &   258,487&0    &   \checkmark   &  $\times$   &   \checkmark \\
    &Valid.    &   -   &0& 98,374   &   \checkmark   &   \checkmark   &   \checkmark \\
    &Valid.    &   EvRep   &   80,537&0     &   $\times$   &   \checkmark    &   \checkmark \\
    &Valid.    &   FakeIm  &   98,373&0     &  \checkmark   &   $\times$   &   \checkmark \\
    \midrule
    \multirow{8}{*}{\makecell{\textbf{Media-}\\\textbf{Eval}\\ \textbf{2015}\\\textbf{(ME)}}}& Train   &   -    &5,245&3,426    &   \checkmark    &   \checkmark    &   \checkmark \\
    & Train    &   EvtRep&2,923&0    &   $\times$  &   \checkmark   &   \checkmark \\
    & Train    &   FakeIm  &3,425&0     &   \checkmark   &   $\times$  &   \checkmark \\
    &Valid.    &   -   &   1,398&  1,479  &   \checkmark    &   \checkmark    &   \checkmark \\
    &Valid.    &   EvtRep  &   0   &   1,267     &   $\times$   &   \checkmark    &   \checkmark \\
    &Valid.    &   FakeIm &   1,479 & 0     &   \checkmark   &   $\times$   &  \checkmark \\
    \cmidrule{2-8}
    &   Test  &   -    & 717& 1,215  &  \checkmark    &   \checkmark    &   \checkmark \\
    \bottomrule
\end{tabularx}
\end{table}
\subsubsection{Manipulation Techniques}
\label{sec:manipulation_techniques}
%
%

To test model generalization, we selected 100 posts from \textit{ME} 2015 labeled as \textit{real} and applied two types of text and image manipulation techniques to generate new test sets. Figure~\ref{Fig: Fig1} depicts an example of each manipulation.
\paragraph{\textbf{Ev}en\textbf{t} \textbf{Rep}lacement}
We randomly replace events that are already provided as meta information in the \textit{ME} dataset with other events from the dataset.
This induces false information, which changes the ground truth of all samples from \textit{real} to \textit{fake}.

\begin{table*}[t]
\small
\centering
\caption{Accuracy~($Acc$) as well as number of samples predicted as fake~($N_{F}$) and real~($N_{R}$) for different models and test data manipulations~(number of fake / real ground-truth samples). Models denoted with~$\ddag$ are solely trained on ME 2015. Note that models with * are reproduced and that VNME-Ens is an ensemble of MLP-CLIP models trained on VNME.}
\label{table:manipulation}
\setlength{\tabcolsep}{4.5pt} 
\begin{tabularx}{1.0\linewidth}{Xccccccccccccccccc}
    \toprule
     \multirow{3}{*}{\bf{Method}} & \bf{ME 2015} & \multicolumn{3}{c}{\bf{Original}} &
    \multicolumn{3}{c}{\bf{FakeIm}} &
    \multicolumn{3}{c}{\bf{RealIm}}   &
    \multicolumn{3}{c}{\bf{EvtRep}}  &
    \multicolumn{3}{c}{\bf{EvtRem}}  &
    \bf{Total}\\
    
     & 
    {\bf{717 / 1,215}}
    &
    \multicolumn{3}{c}{\bf{0 / 100}} &
    \multicolumn{3}{c}{\bf{100 / 0}} &
    \multicolumn{3}{c}{\bf{0 / 100}} &
    \multicolumn{3}{c}{\bf{100 / 0}} &
    \multicolumn{3}{c}{\bf{6 / 94}} &
    \bf{206 / 294}
     \\
    \cmidrule(lr){3-5}
    \cmidrule(lr){6-8}
    \cmidrule(lr){9-11}
    \cmidrule(lr){12-14}
    \cmidrule(lr){15-17}
    
     & 
    \bf{$Acc$} & 
    \bf{$N_{F}$} & 
    \bf{$N_{R}$} & 
    \bf{$Acc$}    &
    \bf{$N_{F}$} &
    \bf{$N_{R}$} &
    \bf{$Acc$}    &
    \bf{$N_{F}$} &
    \bf{$N_{R}$} &
    \bf{$Acc$}    &
    \bf{$N_{F}$} &
    \bf{$N_{R}$} &
    \bf{$Acc$}    &
    \bf{$N_{F}$} &
    \bf{$N_{R}$}    &
    \bf{$Acc$} &
    \bf{$Acc$}
    \\
    \midrule

     BDANN*, \ddag   &  0.76  &  16 &  84  &    0.84    &   12  &  88   &    0.12  & 15 &    85     &   0.85  &    19   &    81     &   0.19  & 17  &     83    &    0.77   &   0.55\\
     Spotfake*, \ddag   &   0.84   &  37    &    63 &    0.63 & 30  & 70    &   0.30  & 18  &    82 &    0.82 & 37  &    63 &    0.37  &   37   &    63 &    0.61 &   0.54    \\
     BERT-ResNet, \ddag   &  0.87  &   28    &   72     &   0.72    &  25   &   75  &   0.25  & 21  &   79  &   0.79  &  28 &   72  &  0.28  &   28   &   72  &    0.68 &   0.54  \\
     CLIP-MMBT, \ddag   &  0.75  &  3   &    97  & \bf{0.97}  &   10   &    90     &   0.10   & 2 &   98  &    0.98  &   4   &  96    &   0.04 &      4  &  96   &    \bf{0.90} &  0.59\\
     \hline
     MLP-CLIP, \ddag &  \bf{0.93}   &  27    &   73   &  0.73 & 40 & 60 &  0.40  & 31 & 69     &   0.69   &   51    & 49 &    0.51  &     39  &  61     & 0.41 &  0.54\\
     
    \textbullet\;VNME-Img   &  0.69   &   3   &  97   &  \bf{0.97}     &   90  & 10    &  0.90  &  5  & 95    &   0.95   &    24    &   76  &    0.24  &     16    &  84    &      0.80  &   0.77\\
    \textbullet\;VNME-Evt   &  0.70   &   6   &   94   &  0.94    &    20   &    80 &  0.20 &    19  &  81    &  0.81   &     75   &  25  &    \textbf{0.75}  &   47    & 53    & 0.51 &   0.64\\
    \textbullet\;VNME-All   &  0.68   &  21   &   79   &  0.79 & 100 & 0 & \bf{1.00}  & 0 & 100   &  \bf{1.00}   &  61 &     39 &   0.61   &  38   &  62   & 0.60 &   0.80\\
    \hdashline
     \textbullet\;VNME-Ens   &  0.70  &   6   &  94   &  0.94  & 100 & 0 &  \bf{1.00}  &   3 & 97    &   0.97   & 62 &  38 &    0.62   &  35   &  65   & 0.63 &  \bf{0.83}\\
    \bottomrule
\end{tabularx}
\end{table*}

\paragraph{\textbf{Ev}en\textbf{t} \textbf{Rem}oval:} Starting from the same initial set, we automatically remove all events from the text. As the ground truth can be both \textit{real} or \textit{fake}, 
one expert with computer science background manually annotated the validity of all 100 samples with removed events and decided on the label \textit{real} or \textit{fake} as follows. The sample remains \textit{real} if 
(1)~the text is still in line with the image, and (2)~the meaning of text has not changed. Otherwise, it is annotated as \textit{fake}. 



\paragraph{Replacement with \textbf{Fake} \textbf{Im}age:} 
We randomly replace images with other images depicting a different event in the test set. Since this produces semantically uncorrelated image-text pairs, the ground truth for these samples is \textit{fake}.


\paragraph{Replacement with \textbf{Real} \textbf{Im}age} 
We replace the input image with 
another similar image from the test set. 
We manually check the chosen image to ensure it is from same event with similar content. Thus, all samples remain \textit{real} after the manipulation. 

\subsubsection{Training Strategies for Improved Generalization}
The \textit{ME} data\-set contains only few real events and limited number of unique images. Thus, 
we extend the dataset using real images and their associated captions from \textit{VisualNews~(VN)}~\cite{DBLP:VN} since it includes much more samples from many domains, topics, and events.
To create \textit{fake} samples, we apply the aforementioned data manipulation techniques as indicated in Table~\ref{tab:VN_data}. 
Considering that events are not provided in \textit{VN}, we apply \textit{spaCy}~\cite{spacy} for named entity recognition to detect named entities and their types. Each entity is replaced randomly with another entity of the same type to realize \textit{Event Replacement~(EvtRep)}. All samples without any detected named entities have been removed since they cannot manipulated.
%

We train our best model~(\textit{MLP-CLIP}) using training data with different manipulations to evaluate their impact. 
As indicated in Table~\ref{tab:VN_data}, the \textit{MLP-CLIP} model~(\textit{VNME-All}) is trained with all training splits, while \textit{MLP-CLIP}~(\textit{VNME-Evt}) and \textit{MLP-CLIP}~(\textit{VNME-Img}) models are trained with the original data and splits manipulated with \textit{Event Replacement} and \textit{Fake Image Replacement}. 
Furthermore, considering that ensemble models tend to have more reliable and better outcome, we evaluate the \textit{MLP-CLIP~(VNME-Ens)} model that combines the outputs of the previous models by majority voting. 
The idea is to utilize all three models which are trained by particular manipulation techniques to reduce the overall error rate.
\subsubsection{Results}
Table~\ref{table:manipulation} shows a comparison of the original and manipulated sets. 
The performance of the models trained exclusively on ME drastically decreases for all manipulated test variants except \textit{RealIm} where the ground truth remains \textit{real}. For the remaining test sets, the ground truth changes to \textit{fake} in most cases, but models tend to predict the same labels for the original and manipulated samples indicating that they are not sensitive to these manipulations~(Figure~\ref{Fig: Fig1}). 
\textit{MLP-CLIP} is the only model that changed predictions towards the correct label. 
However, its performance still drops compared to original test set especially on manipulated sets~(\textit{FakeIm, EvtRep}) where the ground truth changes to \textit{fake}.
%
%

In contrast, our proposed \textit{MLP-CLIP} models that are trained with additional data show significantly better results. 
%
As expected, models trained with datasets including text~(\textit{VNME-Evt}) or image manipulations~(\textit{VNME-Img}) show improved results for the respective modifications.
We conclude that a modality-specific data manipulation technique in the training process improves the 
model generalization and robustness for the respective modality. 
For example, for \textit{MLP-CLIP~(VNME-Img)}, a significant number of samples which were originally predicted as \textit{real} are then predicted as \textit{fake} for \textit{FakeIm} test set.
The model that is trained with all modifications~(\textit{VNME-All}) provides the best overall performance among single models. 
It is only outperformed by the ensemble model~\mbox{\textit{MLP-CLIP~(VNME-Ens)}}, 
which shows that combining models for certain manipulations can improve the performance compared to baselines using a single model, i.e., \textit{Spotfake}~\cite{DBLP:Spotfake}~(Figure~\ref{Fig: Fig1}). While it does not have the best results for each manipulated set, it provides the best overall performance averaged over all test sets which makes it the most reliable model for fake news detection. Overall, the results confirm that more diverse data and the application of data manipulation techniques in the training process improves model generalization.


\section{Conclusion}
\label{sec:conclusion}
In this paper, we have studied the problem of multimodal fake news detection and proposed three multimodal models using state-of-the-art transformer architectures that outperformed baselines on the \textit{2015} and \textit{2016} \textit{MediaEval Benchmark} for the \textit{Verifying Multimedia Use} task.
Furthermore, we have applied text and image manipulation techniques to create more diverse test scenarios, which demonstrated that current approaches tend to overfit and lack generalization due to the limited size and variations of the training data.
We have proposed a solution for this issue by adding more training data from different topics and by applying data manipulation techniques to create diverse samples. 
The models trained on these datasets have shown better generalization capabilities. 
%
%
%
%
In future work, we will explore different kinds of manipulation techniques, fusion strategies to combine different models, and test our approach on larger annotated test sets with more diverse topics. 
%

\begin{acks}
This work was funded by European Union’s Horizon 2020 research and innovation program under the Marie Skłodowska-Curie grant agreement no. 812997 (CLEOPATRA ITN), and by the German Federal Ministry of Education and Research (BMBF, FakeNarratives project, no. 16KIS1517).
\end{acks}

\bibliographystyle{ACM-Reference-Format}
\bibliography{sample-base}

\end{document}